\pdfoutput=1

\documentclass[11pt]{article}

\usepackage{acl}
\usepackage{graphicx}
\graphicspath{ {./narrative_eye_tracking_analysis/images} }
\usepackage{times}
\usepackage{latexsym}
\usepackage{hyperref}
\usepackage[T1]{fontenc}

\usepackage[utf8]{inputenc}

\usepackage{microtype}
\usepackage{float}
\usepackage{ listings }
\usepackage[frozencache=true,cachedir=.]{minted} 
\usepackage{enumitem}
\setlist{leftmargin=5mm}
\setlist[itemize]{noitemsep}
\title{An Analysis of Reader Engagement in Literary Fiction 
\\through Eye Tracking and Linguistic Features}

\usepackage{url}
\usepackage{hyperref}
\usepackage{array}
\newcolumntype{P}[1]{>{\centering\arraybackslash}p{#1}}

\author{Rose Neis \\
 University of Minnesota  \\
 \texttt{neis@umn.edu}
 \And
 Karin de Langis \\
 University of Minnesota\\
 \texttt{dento019@umn.edu} \\
 \AND
 Zae Myung Kim \\
 University of Minnesota\\
 \texttt{kim01756@umn.edu}
 \And
 Dongyeop Kang \\
 University of Minnesota\\
 \texttt{dongyeop@umn.edu} \\
 }

\begin{document}
\pagenumbering{arabic}
\maketitle

\begin{abstract}
Capturing readers' engagement in fiction is a challenging but important aspect of narrative understanding. In this study, we collected 23 readers’ reactions to 2 short stories through eye tracking, sentence-level annotations, and an overall engagement scale survey. We analyzed the significance of various qualities of the text in predicting how engaging a reader is likely to find it. As enjoyment of fiction is highly contextual, we also investigated individual differences in our data. Furthering our understanding of what captivates readers in fiction will help better inform models used in creative narrative generation and collaborative writing tools.
The interactive demo is available here\footnote{\url{https://bookdown.org/bishop_pilot/acldemo2/ACLDemo.html}}.
\end{abstract}

\section{Introduction}
The question of reader engagement in fiction has been studied in the psychology field for decades, with some of the foundational theoretical work from \citet{gerrig_1993} on Transportation Theory paving the way for more recent theoretical frameworks and experimental setups, notably the work by \citet{Green2004} and \citet{busselle2009}.

However, as \citet{jacobs_2015} emphasized in his article on the neurocognitive science of literary reading, the samples normally collected are small and not enough to compensate for individual differences in reading patterns due to reader context and other situational factors.
In order to help close the experimental gap, one contribution of this study is to provide a data set of reader reactions to natural stories, which Jacobs refers to as ``hot'' experimental research. This data, along with the extraction of linguistic features, allows us to test theories around reader engagement and discover which textual qualities have the most impact.

\begin{table*}[h]
\centering
\begin{tabular}{@{}c|P{1cm}|P{2cm}|P{2cm}|P{2cm}|P{2cm}@{}}
\hline
& Ours & \citet{kunze2015} & \citet{Magyari2020} & \citet{HSU201596} & \citet{Maslej2019TheTF} \\
\hline
\multicolumn{6}{@{}l}{\textbf{Data gathered}}\\\hline
Eye tracking & x & x & x &  &  \\\hline
Saccade angle &  & x & x &  & \\\hline
fMRI &  &  &  & x & \\\hline
Engagement survey & x & x & x &  & x\\\hline
Engagement annotation & x &  &  & x & \\\hline
\multicolumn{6}{@{}l}{\textbf{Textual features extracted}}\\\hline
Emotional arc & x &  &  &  & \\\hline
Lexical categories & x &  &  & x & x\\\hline
Description category &  &  & x &  & \\\hline

\end{tabular}
\caption{Comparison between our study and other similar experiments.}
\label{tab:first}
\end{table*}

In our study, we have the following research questions:
\begin{itemize}
    \item \textbf{RQ1: Does absorption in a story lead to longer dwell times?}\label{subsection:rq1} To answer this question, we looked at how well the different annotations correlated with dwell time to see if there is a relationship between dwell time and different modes of reading -- one being immersed and the other more reflective. We also looked at whether linguistic features of the text related to a more affective reading mode led to higher dwell times as Jacobs predicts.
    \item \textbf{RQ2: How much is engagement dependent on reader context vs. linguistic features?}\label{subsection:rq2} In order to address this question, we evaluated how well the features we extracted could predict whether a sentence was highlighted by readers.
    \item \textbf{RQ3: Are dwell time patterns consistent across readers?}\label{subsection:rq3}
    We scaled dwell times per participant and evaluated the pattern over the story to see if dwell times increased and decreased in the same areas of the story for different readers.

\end{itemize}

With respect to \hyperref[subsection:rq1]{RQ1}, our findings indicated that negatively-valenced, concrete sentences had higher dwell times. No relationship was found between the highlights and dwell times. This may be due to the fact that the highlighting data is sparse. For \hyperref[subsection:rq2]{RQ2}, we found that features such as valence, sentiment, and emotion were significant across readers, although the reader context accounted for much of the variance in highlighting annotations. Regarding \hyperref[subsection:rq3]{RQ3}, there was a high amount of variance between readers for dwell time. However, once dwell times were individually scaled, we could see some consistency in their patterns, particularly when looking only at highly engaged readers.

For future studies, a modified highlighting exercise in which participants must select a category for each sentence --- including none --- could result in less sparse annotation data. A more complete annotation of the story text would allow us to explore the connection between dwell time and different modes of engagement. As new methods are created for representing complex features of stories, such as character relationships and story tension, data sets like ours can be used to find more meaningful relationships between the story text and how engaging it is.

\section{Related Work}
In his model for the neurocognitive poetics of literary reading, \citet{jacobs_2015} proposed two modes of reading: one fast track --- ``immersion'' and one slow --- ``aesthetic trajectory''. The former is proposed to be brought on by things like familiarity, suspense, sympathy, and vicarious hope; whereas the latter is a more reflective and connected mode brought on by aesthetic appreciation, more complex emotion, and unfamiliar situations. We used this framework to inform what variables we expected to have an impact on dwell time.

 \citet{busselle2009} conducted a series of studies to narrow down the salient aspects of reader engagement and created a general media engagement scale. The aspects they defined are narrative understanding, attentional focus, emotional engagement, and narrative presence, and the scale they created include questions related to those aspects. We adapted this scale for written narrative to gauge overall interest in the stories used in our study.
In addition, in order to obtain more granular information, we used these aspects to design an annotation task that would provide sentence-level feedback. Using visualizations and linear mixed effect models, we explored textual features that had an impact on engagement and dwell time across readers.
There have been several other eye tracking as well as fMRI studies in the area of reader engagement (a few are shown in \autoref{tab:first}). One 13-participant study showed that words in enactive passages had on average longer fixation durations and dwell times \citep{Magyari2020}. Based on survey responses, the authors hypothesized that in the enactive texts, the ease of imagery contributes to greater involvement in imagination and results in an overall slower reading speed.
\citet{HSU201596} conducted an fMRI study and found valence and arousal scores as good predictors of overall emotional experience of the reader.

\begin{figure*}[t]
\centering
  \includegraphics[width=1\textwidth,trim={2cm 0 0 2cm}]{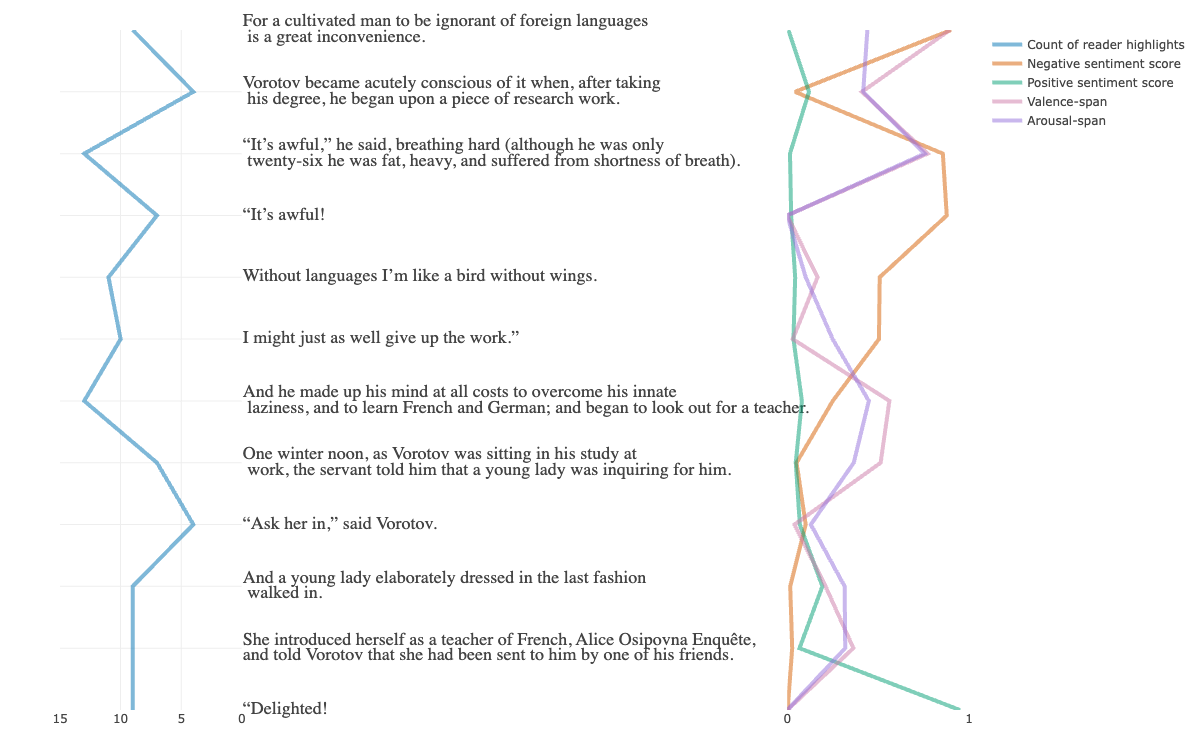}
  \caption{Engagement highlight counts (left) and linguistic feature scores (right) for Expensive Lessons\label{fig:example}. More examples and interactive demos are available in \url{https://bookdown.org/bishop_pilot/acldemo2/ACLDemo.html}}
\end{figure*}

\section{Methods}

\paragraph{Participant study design}
The study asked 31 English speakers (17 female, 11 male, 3 other, average age: 26) to read two short stories by Anton Chekhov\footnote{\hyperlink{https://www.gutenberg.org/cache/epub/13505/pg13505-images.html}{``Expensive Lessons''} and \hyperlink{https://www.gutenberg.org/cache/epub/1732/pg1732-images.html}{``Schoolmistress''}} while their eyes were tracked, and then answer an engagement scale survey:

\begin{itemize}
  \item I was curious about what would happen next. (+)
  \item The story affected me emotionally. (+)
  \item While reading my body was in the room, but my mind was inside the world created by the story. (+)
  \item At times while reading, I wanted to know what the writer's intentions were. (+)
  \item While reading, when a main character succeeded, I felt happy, and when they suffered in some way, I felt sad. (+)
  \item The characters were alive in my imagination. (+)
  \item I found my mind wandering while reading the story. (-)
  \item I could vividly imagine the scenes in the story. (+)
  \item At points, I had a hard time making sense of what was going on in the story (-)
\end{itemize}

After reading through both stories, they completed a highlighting exercise where they highlighted areas according to the following categories:

\begin{itemize}
  \item \textit{Present}: Able to vividly picture the scene in the story
  \item \textit{Confused}
  \item \textit{Curious}: Curious about what will happen next
  \item \textit{Connected}: Connected to the character; able to identify with them or feel their emotions
  \item \textit{Other}: Enjoyed it for a different reason
\end{itemize}

\paragraph{Eye-tracking data}

Due to calibration issues, 8 samples were discarded, leaving 23 (13 female, 8 male, 2 other, average age: 28, std.: 10). See \autoref{tab:fourth} for more details on the participants.
The eye tracking results were drift corrected and interest area reports were exported using words as interest areas. Outliers for dwell time were removed using the inner quartile range method (1.7\% of the data). The data was aggregated to the sentence level and dwell time values were normalized by sentence character count. To handle missing data, null values for the eye tracking features were filled with the average of the 5 nearest sentences (5.7\% of all sentences read across participants). Dwell times were then scaled individually per participant using min-max scaling. This allowed each participant's dwell time patterns to be preserved when scaling.

\begin{figure*}[htp]
\centering
  \includegraphics[width=0.9\textwidth]{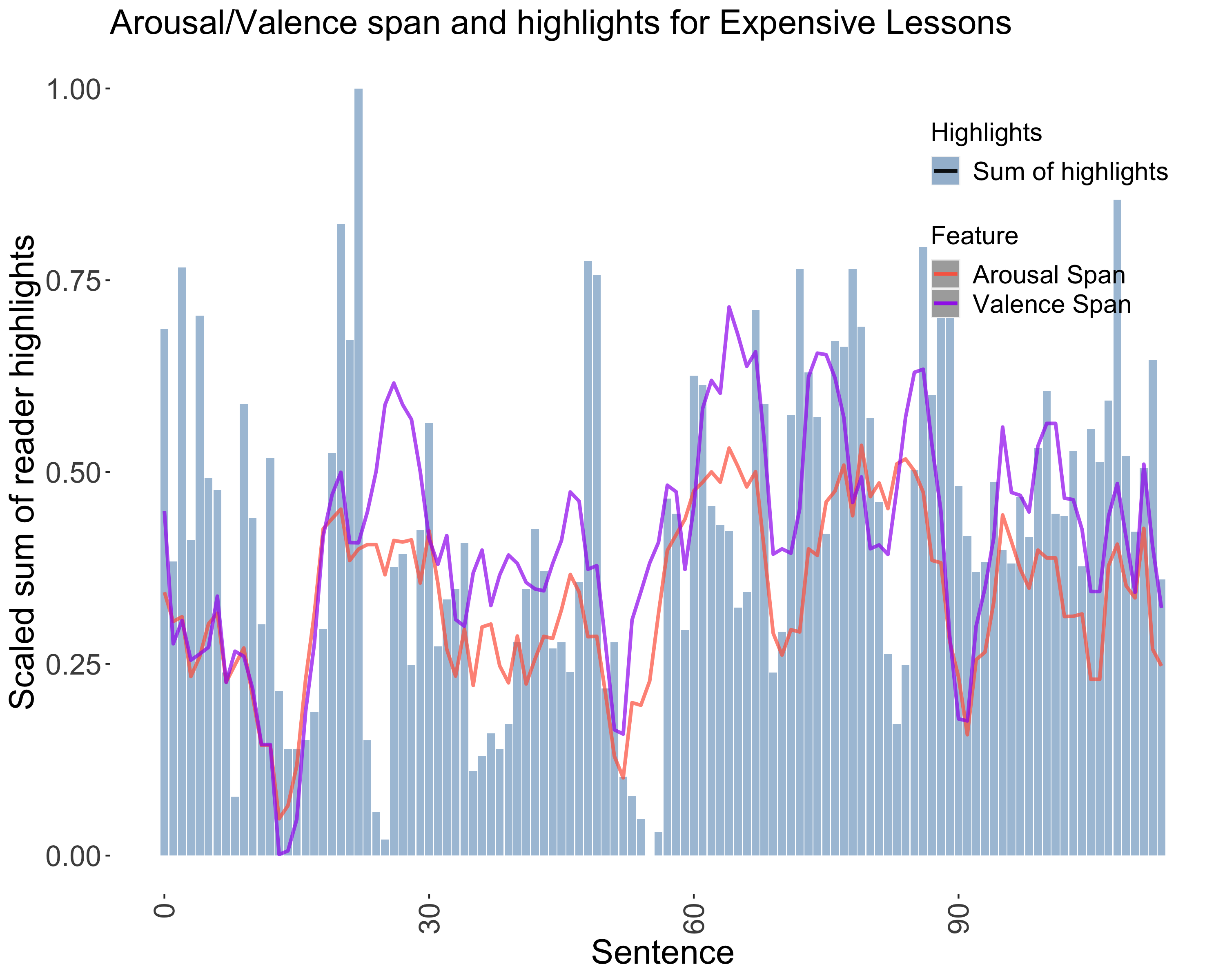}
  \\
  \includegraphics[width=0.9\textwidth]{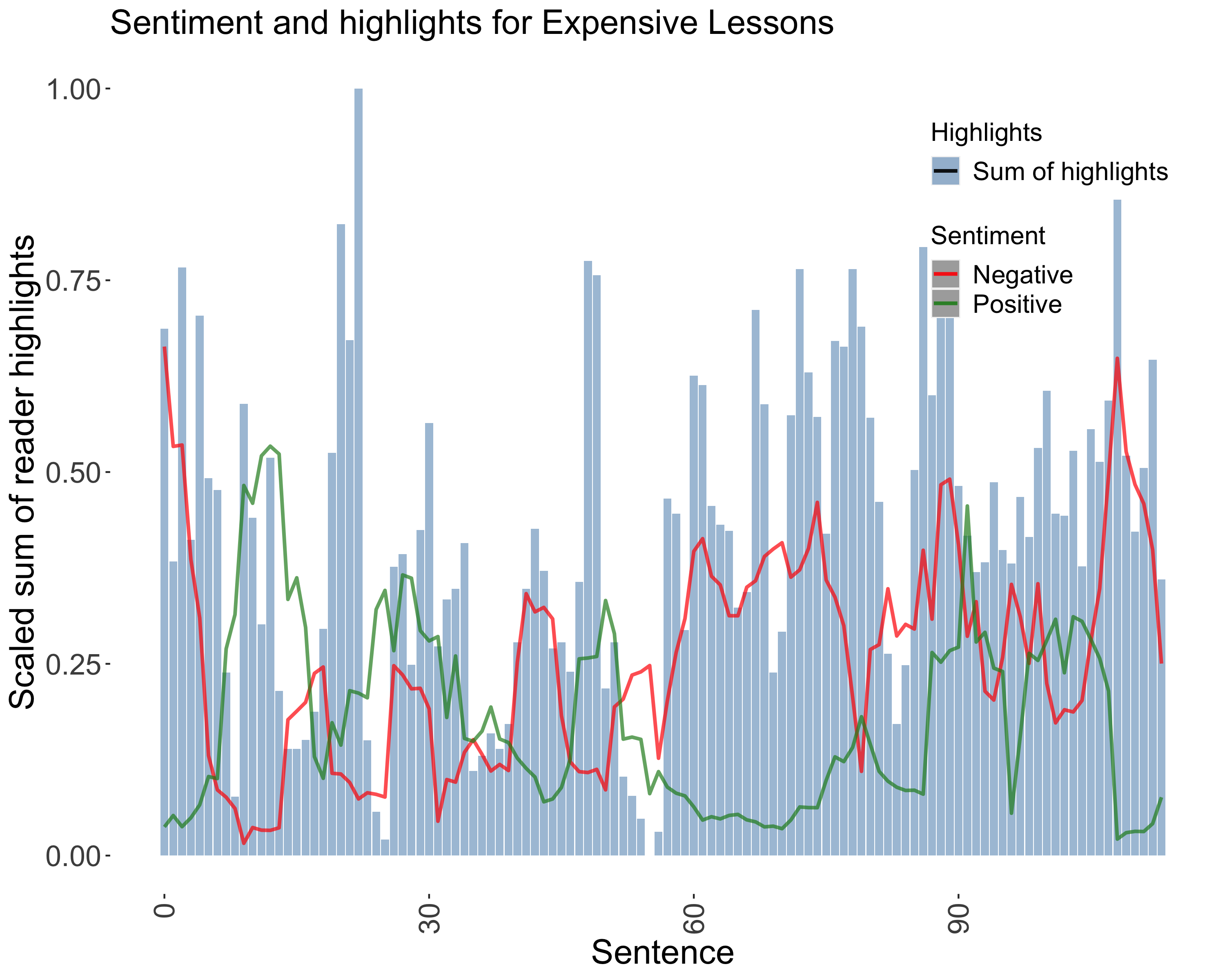}
  \caption{Highlights and features.}
\end{figure*}

\paragraph{Linguistic and discourse features}
We extracted the following features from the stories to create sentence-level predictors: negative and positive sentiment scores using the RoBERTa sentiment base model, emotion categories using the DistilRoBERTa emotion base model\footnote{\hyperlink{https://huggingface.co/cardiffnlp/twitter-roberta-base-sentiment}{RoBERTa sentiment model} and \hyperlink{https://huggingface.co/j-hartmann/emotion-english-distilroberta-base}{DistilRoBERTa emotion model}}, concreteness scores from the \citet{brysbaert2014} corpus, valence and arousal from the NRC-VAD corpus \citep{valence_arousal_dominance_2013}, word frequency from the subtlex corpus \citep{Brysbaert_2015}, and average word length. Emotions extracted were based on the basic emotions described by \citet{ekman2011meant} plus a neutral category: anger, disgust, fear, joy, neutral, sadness, surprise. Sentence level scores for concreteness, valence, arousal, and word frequency were obtained by using the scores of each lemma and then computing the mean and difference between minimum and maximum scores. To obtain lemmas, we used the BookNLP \footnote{\hyperlink{https://github.com/booknlp/booknlp}{BookNLP}} code package. All feature scores used in our models are scaled to $[0, 1]$.

As these sentence-level features can have high variability on their own, we performed low-pass filtering by Fourier transformation on sliding windows of ten sentences. As a result, we were able to filter out extreme features and smoothly track the patterns of features that persist over a longer context.

\paragraph{Limitations}
There are a few issues with the data that should be mentioned. Since the participants were asked to read two stories in a row, it is best to make sure there is a balance in which story is read first. However, due to poor tracking of reading order, our data ended up with a skew towards one story (Expensive Lessons: 16, Schoolmistress: 7), which may affect level of attention for the second story. 

In addition, the stories did not receive high scores on average in the engagement survey. On a scale from 0-4, Expensive Lessons got an average of 2.09 and Schoolmistress averaged 1.92. Ideally, stories used for such studies should be more widely popular in order to make engagement more likely. Perhaps in part due to the low average score, the highlighting data is sparse, making it difficult to find relationships between dwell time and engagement categories.

Finally, although efforts were made to recruit participants from the larger community, a majority of the participants were University students and staff, with a minority from outside the University community. As seen in \autoref{tab:fourth}, this resulted in a skew towards younger, college-educated participants. Observations from this study may not generalize well to other groups.

\section{Results}

Other studies have shown that valence and arousal play an important role in predicting interest in a story \citep{Maslej2019TheTF, HSU201596} and \citet{jacobs_2015} emphasized the importance of affective processes in his framework. In order to determine the importance of these values for our data, we used linear mixed model analysis. Using lme4 \citep{lme4} and lmerTest \citep{lmertest}, we fit predictions of the proportion of the sentence highlighted and dwell time, with random effects of participant (n=23) and story (n=2). Variables were tested for collinearity using the variance inflation factor (VIF) method outlined by \citet{zuur2010}, and no variables exceeded the recommended threshold of 3. Observations and fixed effects are on a $[0,1]$ scale. See \autoref{sec:appendix2} for exact model definitions.

\subsection{Predicting engagement highlights}

\begin{table}[h]
  \begin{tabular}{lrrrr}
  \hline
  & Slope & $Pr(>|t|)$ & Sig. & VIF \\
  \hline
  (Intercept) & -0.05 & 0.49 & & \\
  char. ct. & 0.16 & < 0.001 & $\ast\ast\ast$ & 2.19\\
  word freq. & 0.07 & 0.08 & . & 1.23\\
  positive & 0.03 & 0.09 & . & 1.58\\
  negative & 0.09 & < 0.001 & $\ast\ast\ast$ & 1.73\\
  concrete & 0.02 & 0.15 & & 1.24\\
  valence & 0.11 & 0.011 & $\ast$ & 1.39\\
  arousal & -0.02 & 0.68 & & 1.11\\
  val.-span & 0.11 & < 0.001 & $\ast\ast\ast$ & 2.75\\
  ar.-span & 0.11 & < 0.001 & $\ast\ast\ast$ & 2.61\\
  surprise & 0.08 & 0.001 & $\ast\ast$ & 1.11\\
  disgust & 0.03 & 0.059 & . & 1.15\\
  \hline
  \end{tabular}
  \caption{Fixed Effects: predicting highlights}
  \label{tab:second}
  \end{table}

We fit a model for predicting the proportion of a sentence highlighted by a reader in order to see how significant the textual features were across readers to address \hyperref[subsection:rq2]{RQ2}. \autoref{tab:second} shows major results in predicting annotated highlights with different linguistic and discourse features. 

Our results support a significance of valence mean (p=0.01), similar to \citet{HSU201596}. Unlike in other studies, we found that arousal mean had no significance (p=0.686).  However, similar to \citet{HSU201596}, valence-span --- the difference between valence max and valence min (p<0.001) and arousal-span --- the difference between arousal max and arousal min (p<0.001) were significant. The positive slope for both (0.1) suggests that the reader was more engaged in sentences with a higher range of valence and arousal.

Of the emotion categories (i.e. anger, disgust, fear, joy, neutral, sadness, surprise), surprise was found to be a significant effect (p=0.001) with a positive slope (0.08). Other features that had an impact were negative sentiment score (p<0.001) and character count (p<0.001). The positive slope for negative sentiment (0.09) partially align with the \citet{Maslej2019TheTF} study, where negative emotion predicted higher story ratings, although unlike their findings, there was no relationship between concreteness and engagement.

When including random effects that model individual participants, the model explains 23\% of the variance; without these effects the explained variance drops to 3.7\%. So, with respect to \hyperref[subsection:rq2]{RQ2}, the reader context is important in elucidating the relationships of the fixed effects with engagement.

Since the proportion is bounded between 0 and 1, the model residuals are not normally distributed. We therefore also fit a generalized mixed model with a binomial distribution, with the observed outcome a binary variable representing whether or not the sentence had any highlighting. \autoref{tab:regression_results} shows largely the same results, except that word frequency and positive sentiment are not significant when predicting the binary outcome.

\subsection{Predicting eye movement dwell time}

\begin{table}[h]
  \centering
  \begin{tabular}{lrrrr}
  \hline
  & Slope & $Pr(>|t|)$ & Sig. & VIF \\
  \hline
  (Intercept) & 0.10 & < 0.001 & $\ast\ast\ast$ &  \\
  word freq. & 0.18 & < 0.001 & $\ast\ast\ast$ & 1.19\\
  positive & 0.01 & 0.045 & $\ast$ & 1.57\\
  negative & 0.01 & 0.1 & & 1.68\\
  concrete & 0.01 & 0.0002 & $\ast\ast\ast$ & 1.21 \\
  valence & -0.06 & < 0.001 & $\ast\ast\ast$ & 1.37 \\
  arousal & -0.01 & 0.34 & & 1.09\\
  val.-span & -0.02 & 0.0029 & $\ast\ast$ & 2.40\\
  ar.-span & -0.06 & < 0.001 & $\ast\ast\ast$ & 2.24\\
  surprise & -0.03 & < 0.001 & $\ast\ast\ast$ & 1.07\\
  \hline
  \end{tabular}
  \caption{Fixed Effects: predicting dwell time}
  \label{tab:third}
  \end{table}

\begin{figure*}[ht]
  \centering
  \includegraphics[height=8cm]{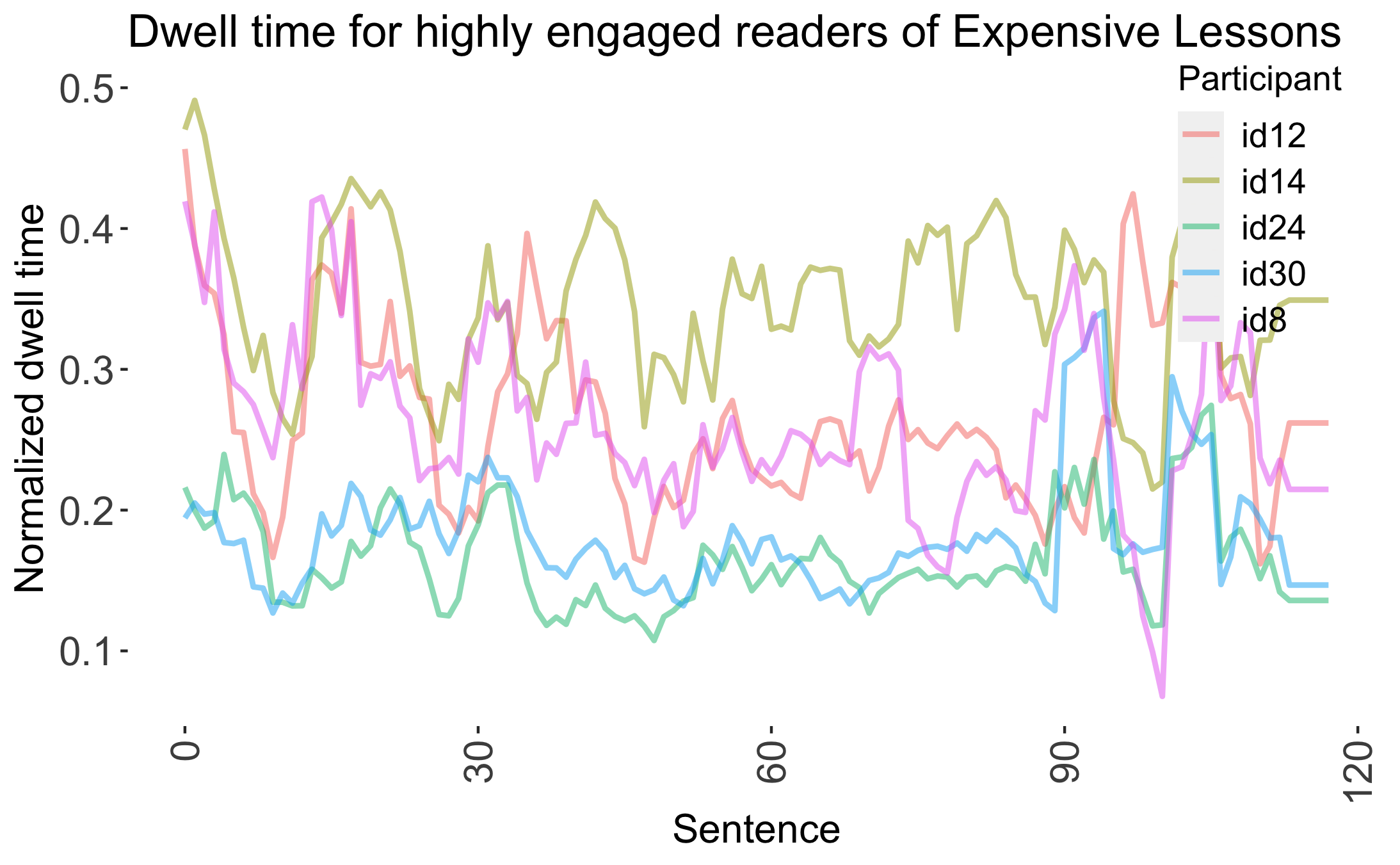}
  \caption{Dwell time for engaged readers}
  \label{fig:fig2}
\end{figure*}

To address \hyperref[subsection:rq1]{RQ1}, we fit a model that predicted dwell time (\autoref{tab:third}). In our findings, valence mean was significant (p<0.001) with a negative slope (-0.06) and arousal mean was not (p=0.349). Valence-span (p=0.0029) and arousal-span (p<0.001) were found to be significant. The negative relationship between valence mean and dwell time supports part of Jacobs' proposed framework, which states that passages that engage our emotions, particularly negative valence, would likely result in higher dwell times. There was no relationship between highlights and dwell time, however, so we were not able to confirm whether the different categories of engagement correlated with different modes of reading.

There was also a positive relationship between concreteness and dwell time (p<0.001, slope=0.01). According to the prevailing theory in neuroscience, "words referring to easily perceptible entities coactivate the brain regions involved in the perception of those entities" \citep{brysbaert2014}. This observation may indicate that this leads to longer processing times. So indirectly our observation has some overlap with the findings of \citet{Maslej2019TheTF}, where enactive passages had higher dwell times, although the linguistic features of their study differed.

To evaluate how consistent dwell time patterns were across readers (\hyperref[subsection:rq3]{RQ3}), we examined the dwell time graphs of participants to see if there was a similar pattern. We noticed an especially striking similarity in patterns amongst readers who were highly engaged (see \autoref{fig:fig2}).

Although removing word-level outliers for dwell time improved the skewness of the data, it is still heavily skewed to the left. This resulted in residuals with a fat tail and therefore not perfectly normal. A log transformation improved the normality of the data, but it resulted in less normal residuals. This may impact the reliability of the above results.

\section{Conclusion}

By collecting reader feedback and eye tracking data on literary fiction, we were able to support findings of other studies that emphasized the importance of affective language for reader immersion. Although we found no direct relationship between dwell times and highlighted text, the dwell time model and the highlight model shared some predictors, such as valence and arousal. One possibility to explore for future studies would be to look at whether this overlap is related to two different modes of engagement --- one that leads to higher dwell times and one that leads to lower dwell times. 

However, as mentioned, this exploration would require a more complete annotation. This could be achieved by selecting more engaging stories and modifying the highlighting exercise to require readers to annotate each sentence with a category or select none. Further analysis on our data set could be done by extracting more complex features. This would expand the analysis beyond the lexical level would allow us to find more interesting relationships.

\nocite{Green2004,liwc_22,brysbaert2014,Maslej2019TheTF,green_brock_kaufman_2006,Consoli2018,busselle2009,jacobs2018,jacobs2017,stockwell2002cognitive,HSU201596,jacobs_2015,mak2019,kunze2015,delatorre2019,indico2015,gerrig_1993,Magyari2020,valence_arousal_dominance_2013,Brysbaert_2015,lme4,lmertest,zuur2010,vifs2012,hartmann2022emotionenglish,ekman2011meant,booknlp,barbieri-etal-2020-tweeteval}

\bibliographystyle{acl_natbib}
\bibliography{anthology,custom}

\section{Acknowledgements}
We would like to thank the University of Minnesota Text Group for initial feedback on experiment setup and the Blue Lantern Writing Group for additional feedback.

\appendix

\section{Participants}
\label{sec:appendix1}

\begin{table}[htbp]
\centering
\begin{tabular}{|l|c|}
\hline
\textbf{Category} & \textbf{Count} \\
\hline
Age: 18-24 & 10 \\
Age: 25-40 & 10 \\
Age: 40+ & 3 \\
\hline
Native English speaker & 17 \\
\hline
Speaks English with friends & 23 \\
Speaks English with family & 22 \\
Speaks English at work & 23 \\
\hline
Gender: Female & 13 \\
Gender: Male & 8 \\
Gender: Other & 2 \\
\hline
\end{tabular}
\caption{Participant info (n=23)}
\label{tab:fourth}
\end{table}

\section{Model Definition}
\label{sec:appendix2}

Predicting engagement highlights: 
\begin{minted}[tabsize=2,breaklines]{R}
lmer(proportion_highlighted ~ 1 + norm_dwell_time + character_count_norm + word_freq_avg + positive + negative + concreteness  + valence_avg + arousal_avg + valence_span + arousal_span + surprise + disgust + (1 | participant_id) + (1 | story), data=all_data)
\end{minted}

Predicting dwell time:
\begin{minted}[tabsize=2,breaklines]{R}
lmer(norm_dwell_time ~ 1  + word_freq_avg  + positive + negative + concreteness + valence_avg + arousal_avg +valence_span  + arousal_span + surprise + proportion_highlighted + (1 | participant_id) + (1 | story), data=all_data)
\end{minted}

Predicting engagement highlights --- binary:

\begin{minted}[tabsize=2,breaklines]{R}
glmer(highlight ~ 1 + norm_dwell_time + character_count_norm + word_freq_avg + positive + negative + concreteness  + valence_avg + arousal_avg + valence_span + arousal_span + surprise + disgust + (1 | RECORDING_SESSION_LABEL) + (1 | story), data=all_data, family=binomial)
\end{minted}

\begin{table}[h]
\centering
\begin{tabular}{lrrr}
\hline
Variable              & Estimate & Pr($>|z|$) & Sig  \\
\hline
(Intercept)           & -2.51    & <0.001     & ***  \\
norm\_dt              & 0.28     & 0.279      &      \\
char. count & 2.05     & < 0.001    & ***  \\
word freq.        & 0.17     & 0.569      &      \\
positive              & 0.08     & 0.548      &      \\
negative              & 0.44     & 0.001      & ***  \\
concreteness          & 0.01     & 0.876      &      \\
valence       & 0.88     & 0.006      & **   \\
arousal        & -0.32    & 0.392      &      \\
val.-span        & 0.90     & <0.001     & ***  \\
ar.-span         & 0.78     & <0.001      & ***  \\
surprise              & 0.51     & 0.002      & **   \\
disgust               & 0.19     & 0.049      & *    \\
\hline
\end{tabular}
\caption{Results from binomial engagement highlights model}
\label{tab:regression_results}
\end{table}

\section{Further Experiments}
\label{sec:appendix3}

We asked ChatGPT to highlight the most engaging parts of the beginning of ``Expensive Lessons'', and here was the response:

Highlighted engaging portions: 
\begin{itemize}
    \item The opening sentence immediately grabs the reader's attention by presenting an idea that they may not have considered before.
    \item Vorotov's struggle with his lack of language skills is relatable and adds a sense of realism to the story.
    \item The description of Vorotov's physical appearance and breathing difficulties creates a vivid image of the character and adds depth to his struggle.
    \item The use of dialogue between Vorotov and the servant adds a sense of immediacy to the story and keeps the reader engaged.
    \item The introduction of Alice Osipovna Enquête as a young lady dressed in the latest fashion creates intrigue and mystery around her character. 
    \item The revelation that Alice was sent by one of Vorotov's friends adds a layer of unexpectedness to the story and piques the reader's curiosity.
\end{itemize}

This shows good potential for automatically detecting the most engaging passages to provide an engaging summary of stories.

\end{document}